\newcommand{\CLASSINPUTinnersidemargin}{0.75in}
\newcommand{\CLASSINPUToutersidemargin}{0.75in}
\newcommand{\CLASSINPUTtoptextmargin}{0.75in}
\newcommand{\CLASSINPUTbottomtextmargin}{0.75in}

\documentclass[conference,letterpaper]{IEEEtran}

\IEEEoverridecommandlockouts

\renewcommand{\IEEEtitletopspaceextra}{0.25in}

\usepackage{cite}
\usepackage{amsmath,amssymb,amsfonts}
\usepackage{graphicx}
\usepackage{textcomp}
\usepackage{xcolor}
\usepackage{booktabs}
\usepackage{makecell}
\usepackage{caption}
\usepackage{subcaption}
\usepackage{enumitem}
\usepackage{hyperref}

\setlist{nosep}

\def\BibTeX{{\rm B\kern-.05em{\sc i\kern-.025em b}\kern-.08em
    T\kern-.1667em\lower.7ex\hbox{E}\kern-.125emX}}

\begin{document}

\title{LooperMuscle: Fast and Stable Learning of Humanoid Whole-Body Tracking via Structured Mixture-of-Experts}

\author{\IEEEauthorblockN{Boyi Liu$^{1, 2}$, Qijin Li$^{1}$, Tianqi Yu,$^{1}$, Qinrui Yan,$^{1}$ and Xingxing Zuo$^{3,\ast}$}

\thanks{$^{1}$DeepMirror Inc., Guangzhou, China.}%
\thanks{$^{2}$HKUST, Hongkong SAR, China.}%
\thanks{$^{3}$Mohamed bin Zayed University of Artificial Intelligence (MBZUAI), Abu Dhabi, UAE.}%
\thanks{$^\ast$Corresponding author. (Email: {\tt\small xingxing.zuo@mbzuai.ac.ae})}
}

\maketitle

\begin{abstract}
FastSAC-style methods significantly reduce humanoid motion training time but often suffer from notable performance degradation compared with PPO in whole-body tracking tasks. We target this speed–performance gap by introducing LooperMuscle, a composed expert policy learning framework that restores tracking quality while preserving high training efficiency. LooperMuscle combines a semantically structured mixture-of-experts actor, an expert-aware distributional critic, and contribution-routed replay with deferred curriculum scheduling. These three components form a closed training loop in which expert contributions guide data routing, routed data shape value learning, and value gradients in turn refine expert specialization. Empirically, our approach substantially outperforms vanilla FastSAC in motion tracking accuracy while requiring far less wall-clock time than PPO: where FastSAC trains in about 15 minutes but underperforms, and PPO achieves stronger results but requires about 6 hours, LooperMuscle recovers a substantial fraction of the remaining gap to PPO in roughly 45 minutes of simulation training, delivering practical efficiency for rapid policy iteration. The code will be released to benefit the research community at: \href{https://loopermuscle.github.io/}{https://loopermuscle.github.io}.
\end{abstract}


\section{Introduction}\label{intro}
Reinforcement learning (RL) combined with massively parallel simulation has dramatically reduced the training time for robot control policies, from hours to minutes in many benchmark tasks \cite{rudin2022learning, makoviychuk2021isaac, zakka2025mujoco}. This acceleration is particularly valuable for sim-to-real development of humanoid robots, where the iterative cycle of training, deploying, identifying domain gaps, and retraining must be repeated until the policy is reliable \cite{zhao2020sim, chebotar2019closing}. Recent work on off-policy RL algorithms, notably FastSAC~\cite{seo2025learning} and FastTD3 \cite{seo2025fasttd3}, has pushed this further by enabling humanoid locomotion training in as little as 15 minutes on a single GPU. However, when the task shifts from locomotion to whole-body tracking (WBT) --- where the policy must faithfully reproduce diverse, high-dimensional human motions across all joints --- fast training alone is insufficient. The tracking quality achieved by off-policy methods still falls notably short of on-policy baselines such as PPO \cite{schulman2017proximal}, revealing a persistent speed--performance gap that limits the practical utility of rapid training recipes.

We argue that this gap stems from three interrelated structural limitations. First, monolithic policy networks struggle to handle the heterogeneous dynamics across body parts, forcing a compromise between lower-body stability and upper-body expressiveness. Second, scalar value functions provide only a single aggregate quality estimate, making credit assignment to individual body parts difficult in high-dimensional WBT. Third, uniform experience replay treats all transitions equally, causing well-performing policy components to dominate the gradient signal while under-performing components stagnate. These three limitations reinforce each other, as a poorly structured policy generates imbalanced experience that a coarse critic cannot correctly diagnose.

\begin{figure}[t]
    \centering
    \includegraphics[width=0.96\columnwidth]{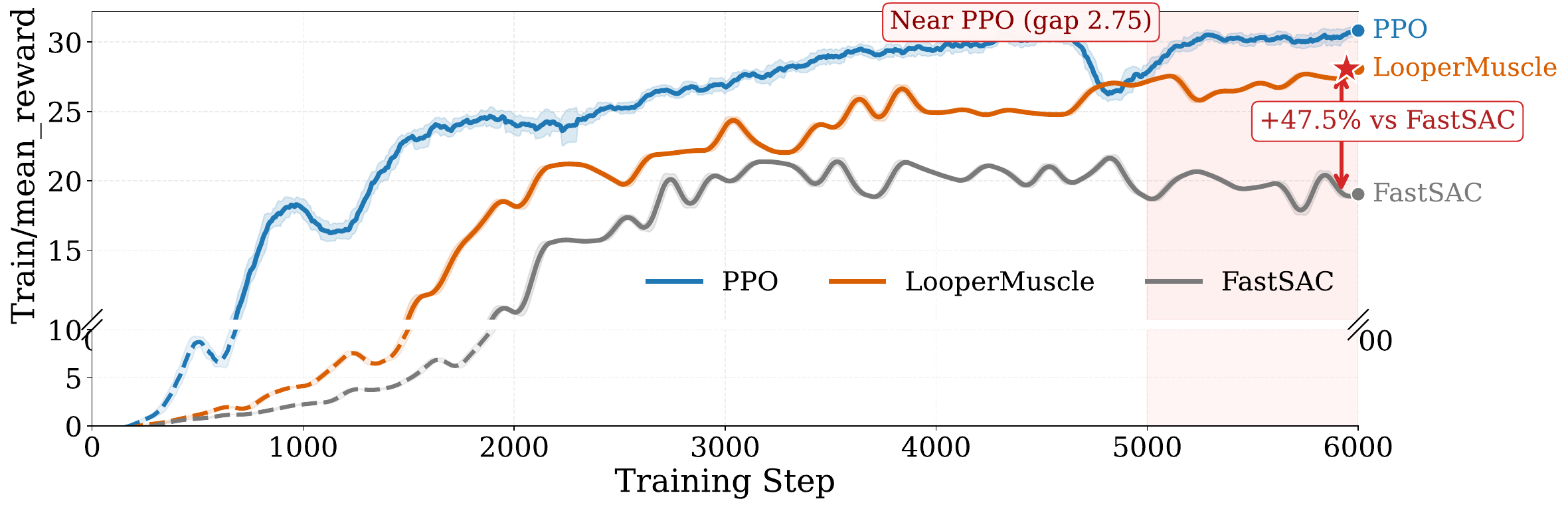}
    \includegraphics[width=0.93\columnwidth]{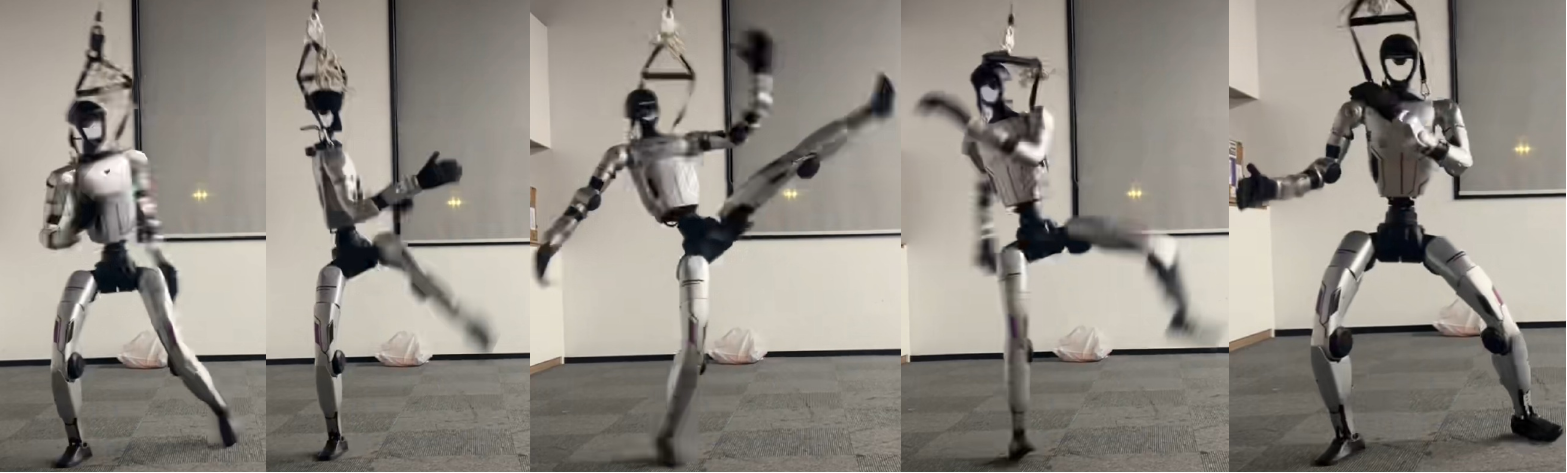}
\caption{\textbf{LooperMuscle overview.} \underline{\textit{Top:}} Training reward curves on a reference motion. LooperMuscle converges in $\sim$45\,min with 47.5\% higher reward than FastSAC and \textbf{$\sim$8$\times$ faster than PPO}. \underline{\textit{Bottom:}} Real-world deployment on Unitree G1 executing a motion sequence from the KungfuBot2 motion library~\cite{han2025kungfubot2}, demonstrating stable and expressive whole-body tracking.}
\label{fig:teaser}
\end{figure}

We introduce LooperMuscle, a composed expert actor-critic framework that addresses all three limitations through a tightly integrated design. The three mechanisms form a closed loop --- expert contributions route data, routed data shapes distributional value estimates, and value gradients refine expert specialization. As illustrated in Fig.~\ref{fig:teaser}, LooperMuscle converges in $\sim 45$\,min with $47.5\%$ higher reward than FastSAC and $\sim 8\times$ faster than PPO, and the trained policy transfers to real hardware for stable whole-body tracking. Our contributions are as follows.

\begin{itemize}[leftmargin=*]
\item \textbf{Multi-expert policy with anti-collapse gating.} We propose a Mixture-of-Experts actor for 29-DoF whole-body tracking, where multiple experts are fused through per-joint-group gating with learnable semantic scaling and output alignment. A KL-divergence-based load-balance regularizer prevents gate collapse, ensuring that weaker experts remain active throughout training.
\item \textbf{Expert-aware distributional critic.} We construct per-expert DVF (Distributional Value Function)~\cite{bellemare2017distributional}, aggregated via gating weights to form the critic output. This makes the value structure isomorphic to the actor's expert structure, enabling finer credit assignment while preserving dual-critic SAC stability.
\item \textbf{Quota-routed replay with deferred scheduling.} We introduce contribution-routed replay that allocates batch slots according to per-expert quotas, combined with progress-dependent deferred buckets that control when difficult samples enter training --- stabilizing early convergence while progressively raising the performance ceiling.
\item \textbf{Empirical validation on humanoid WBT.} On the 40-sequence benchmark, LooperMuscle reaches 72\% of PPO's converged normalized reward and reduces body error by 34\% over FastSAC-MLP (65\% of PPO) in roughly 45 minutes, versus PPO's $\sim$6-hour training; on individual motions it recovers up to $\sim$90\% of PPO's reward (Fig.~\ref{fig:teaser}). Ablations confirm the necessity of each component.
\end{itemize}

\section{Related Work}
\label{sec:related}
\textbf{Humanoid whole-body tracking.}
Physics-based character animation and humanoid motion imitation have a long history. DeepMimic \cite{peng2018deepmimic} established the paradigm of training RL policies to track reference motions through per-joint reward shaping and early termination. Subsequent works extended this to adversarial imitation \cite{peng2021amp}, large-scale motion datasets \cite{luo2023perpetual}, and progressive training curricula \cite{luo2024universal}. More recently, BeyondMimic \cite{liao2025beyondmimic} demonstrated versatile humanoid tracking with lightweight reward structures and diffusion-based policy guidance, while HOVER \cite{he2025hover} and ASAP \cite{he2025asap} achieved agile whole-body skills through careful sim-to-real pipelines. A common thread across these methods is the use of monolithic policy networks that must jointly handle all body parts. In contrast, LooperMuscle decomposes the policy into multiple specialized experts with per-joint-group gating, explicitly addressing the heterogeneous dynamics across body regions.

\textbf{Off-policy RL for humanoid control.}
PPO \cite{schulman2017proximal} remains the dominant algorithm for sim-to-real humanoid learning, partly because on-policy methods scale naturally with massively parallel simulation \cite{rudin2022learning, makoviychuk2021isaac, zakka2025mujoco}. However, recent work has demonstrated that off-policy algorithms can also scale effectively in large-scale regimes while achieving faster wall-clock convergence. Parallel Q-learning \cite{li2023parallel} showed competitive performance with thousands of parallel environments. Raffin \cite{raffin2025sac} and Shukla \cite{shukla2025sac} independently investigated SAC-based training at massive scale. Most notably, Seo et al.\ \cite{seo2025learning, seo2025fasttd3} introduced FastSAC and FastTD3, achieving humanoid locomotion training in 15 minutes and the first sim-to-real deployment of off-policy humanoid controllers. Their recipe employs distributional critics (DVF; \cite{bellemare2017distributional}), observation and layer normalization, and careful hyperparameter tuning. However, these methods use standard single-network policies and do not modify the critic or replay structure to account for multi-expert interactions. LooperMuscle builds directly on the FastSAC recipe but augments it with structured expert decomposition in both the actor and critic, along with contribution-aware data routing.

\textbf{Mixture-of-experts and distributional RL.}
MoE architectures have seen renewed interest in deep learning \cite{shazeer2017outrageously, fedus2022switch}, with load-balance losses widely used to prevent expert collapse. In RL, multi-expert policies have been explored for multi-task and hierarchical settings \cite{yang2020multi, ren2021probabilistic}, but remain rare in high-dimensional humanoid control. Concurrently, KungfuBot2~\cite{han2025kungfubot2} introduces Orthogonal MoE with Gram--Schmidt orthogonalization for versatile whole-body control from 9{,}770 motion sequences under PPO, showing clear gains over monolithic MLPs. On the value side, distributional RL replaces scalar estimates with full return distributions; DVF \cite{bellemare2017distributional} and QR-DQN \cite{dabney2018distributional} demonstrated benefits in discrete domains, and distributional critics have since been adopted in continuous control \cite{li2023parallel, seo2025fasttd3}, though always with a single monolithic head. LooperMuscle targets a complementary objective to KungfuBot2 --- fast per-task training rather than universal versatility --- and combines both directions: a per-joint-group MoE actor paired with per-expert distributional critic heads and contribution-routed replay, a combination not previously explored for humanoid whole-body tracking.
\section{Methodology}
\label{sec:method}

\begin{figure*}[t]
\centering
\includegraphics[width=0.92\textwidth]{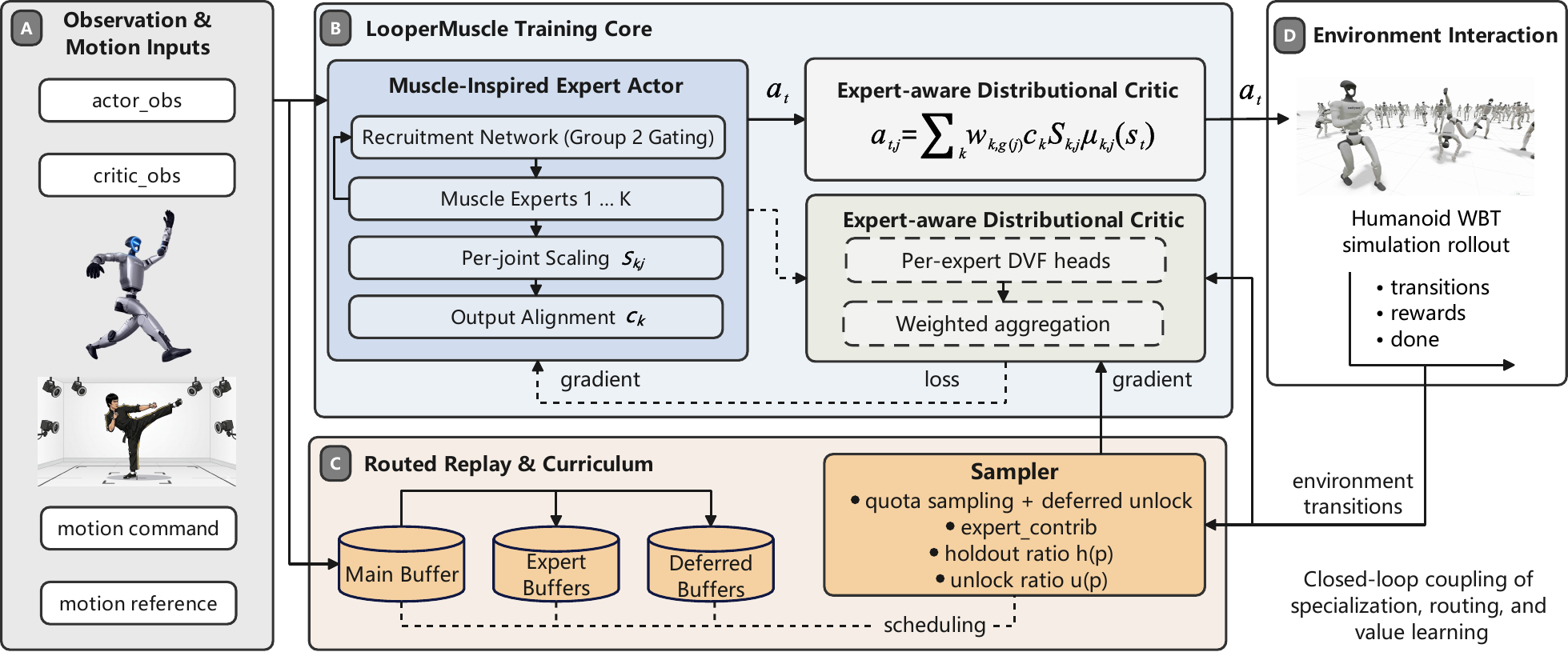}
\caption{\textbf{System overview of LooperMuscle.} (A)~The policy receives proprioceptive observations and motion references. (B)~The Muscle-Inspired Expert Actor decomposes the policy into $K$ experts fused via per-joint-group gating $w_{k,g}$, with per-joint scaling $S_{k,j}$ and output alignment $c_k$. The Expert-Aware Distributional Critic maintains per-expert DVF heads aggregated through the same gating weights. (C)~Routed Replay \& Curriculum stores transitions with expert contribution metadata and samples via quota routing with progress-dependent deferred unlock. (D)~Actions are executed in massively parallel simulation. Together, the three components close the loop among policy structure, data allocation, and value estimation.}
\label{fig:framework}
\end{figure*}

This section presents the LooperMuscle framework. We define the problem setup (Sec.~\ref{sec:setup}), then describe the MoE actor with anti-collapse gating (Sec.~\ref{sec:actor}), expert-aware distributional critic (Sec.~\ref{sec:critic}), quota-routed replay with deferred scheduling (Sec.~\ref{sec:replay}), and closed-loop integration (Sec.~\ref{sec:loop}). An overview is shown in Fig.~\ref{fig:framework}.

\subsection{Problem Setup and Notation}
\label{sec:setup}

We consider whole-body tracking (WBT) for a humanoid robot with $d = 29$ degrees of freedom (DoF). At each timestep $t$, the policy observes a state $\mathbf{s}_t$ comprising proprioceptive measurements (joint positions $\mathbf{q}_t \in \mathbb{R}^{d}$, velocities $\dot{\mathbf{q}}_t \in \mathbb{R}^{d}$, body orientation $\mathbf{r}_t \in \mathbb{R}^{3}$, angular velocity $\boldsymbol{\omega}_t \in \mathbb{R}^{3}$) and motion reference signals (target joint positions $\mathbf{q}^{\text{ref}}_{t} \in \mathbb{R}^{d}$, key-body positions $\mathbf{p}^{\text{ref}}_{t} \in \mathbb{R}^{N_{\text{key}} \times 3}$, target orientation $\mathbf{r}^{\text{ref}}_{t} \in \mathbb{R}^{3}$). The policy outputs $\mathbf{a}_t \in \mathbb{R}^d$ as target joint positions sent to PD controllers.

We partition the $d$ joints into $G$ semantic groups via a mapping $g: \{1,\ldots,d\} \to \{1,\ldots,G\}$. In our implementation, we use $G=2$ groups corresponding to upper-body and lower-body joints, reflecting the distinct dynamic roles of these regions (balance and locomotion vs.\ expressive arm and torso motions).

Our framework builds on Soft Actor-Critic (SAC)~\cite{haarnoja2018soft} with the FastSAC recipe~\cite{seo2025learning} as the base training algorithm. We extend it with a structured multi-expert actor, a distributional critic with per-expert heads, and a contribution-aware replay mechanism described below.

\subsection{Multi-Expert Actor with Anti-Collapse Gating}
\label{sec:actor}

\begin{figure}[t]
\centering
\includegraphics[width=0.92\columnwidth]{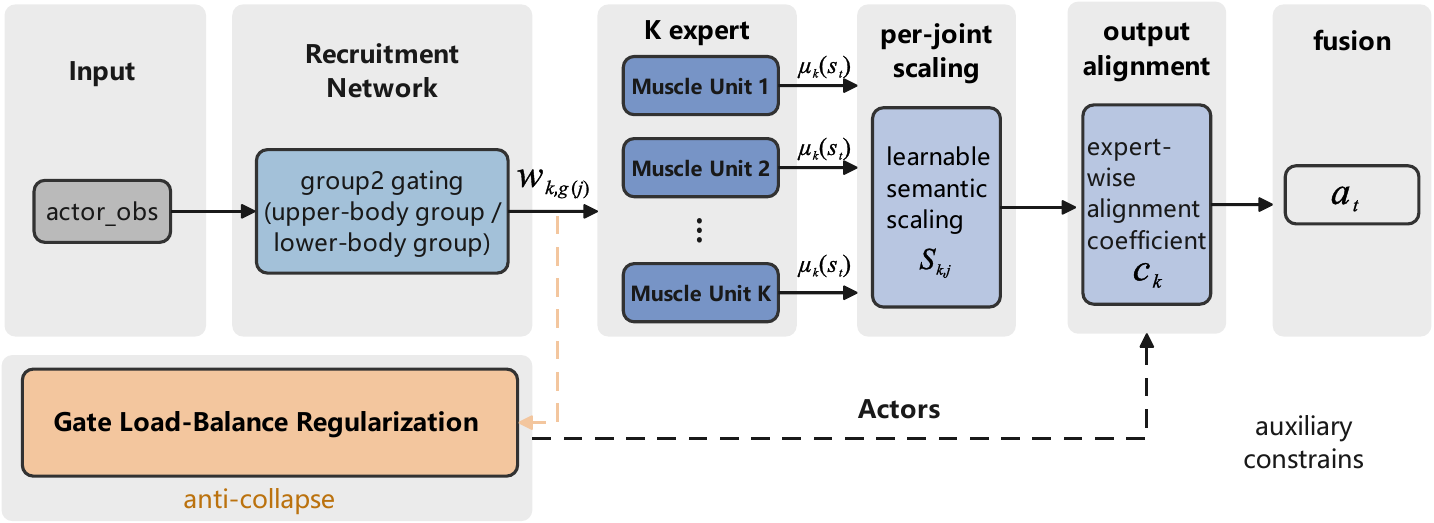}
\caption{\textbf{MoE Actor architecture.} The input observation is processed by a recruitment network that produces per-joint-group gating weights $w_{k,g}(\mathbf{s}_t)$. Each Muscle Unit (expert) outputs actions $\boldsymbol{\mu}_k(\mathbf{s}_t)$, modulated by learnable per-joint scaling $S_{k,j}$ and output alignment $c_k$ before fusion via Eq.~\eqref{eq:fusion}. A KL-based gate load-balance regularizer prevents gate collapse.}
\label{fig:actor}
\end{figure}

A monolithic policy network must compromise between the competing dynamics of different body regions. We decompose the policy into $K$ expert networks fused through learnable per-joint-group gating. The architecture is illustrated in Fig.~\ref{fig:actor}.

\textbf{Expert fusion.}
Each expert $k \in \{1,\ldots,K\}$ produces a per-joint action $\mu_{k,j}(\mathbf{s}_t)$. The final action is:
\begin{equation}
a_{t,j} = \sum_{k=1}^{K} w_{k,g(j)}(\mathbf{s}_t) \; c_k \; S_{k,j} \; \mu_{k,j}(\mathbf{s}_t), \quad j = 1,\ldots,d
\label{eq:fusion}
\end{equation}
where $w_{k,g(j)}(\mathbf{s}_t)$ is the gating weight for expert $k$ in the joint group containing joint $j$ (computed at the \emph{group level}, shared within each group), $S_{k,j}$ is a learnable per-joint scaling factor initialized from the nominal joint range on Unitree G1, and $c_k$ is an output alignment coefficient normalizing magnitudes across experts. This two-level design --- group-level gating for coarse expert recruitment, per-joint scaling for fine-grained modulation --- allows multiple experts to contribute to each joint with state-dependent ratios.

\textbf{Per-joint-group gating.}
Gating weights are computed independently for each group via temperature-scaled softmax:
\begin{equation}
w_{k,g}(\mathbf{s}_t) = \frac{\exp\bigl(\ell_{k,g}(\mathbf{s}_t) / \tau_g\bigr)}{\sum_{k'=1}^{K} \exp\bigl(\ell_{k',g}(\mathbf{s}_t) / \tau_g\bigr)}, \quad \sum_{k=1}^{K} w_{k,g}(\mathbf{s}_t) = 1
\label{eq:gating}
\end{equation}
where $\ell_{k,g}(\mathbf{s}_t)$ are gating logits from a shared recruitment network and $\tau_g > 0$ is a per-group temperature. Lower $\tau_g$ encourages sharper selection (for stability-critical lower-body), while higher $\tau_g$ allows softer blending (for expressive upper-body). This enables upper-body and lower-body to recruit different expert combinations independently.

\textbf{Semantic scaling and output alignment.}
The scaling factors $S_{k,j}$ are initialized proportionally to the angular range $(\bar{q}_j - \underline{q}_j)$ of joint $j$, reflecting heterogeneous action scales: hip joints have large ranges (larger scaling) while wrist joints operate in smaller ranges (smaller scaling). Both $S_{k,j}$ and the alignment coefficients $c_k$ remain learnable during training.

\textbf{Anti-collapse gating regularization.}
To prevent gate collapse, where a single expert monopolizes gating weights, we introduce a load-balance regularizer:
\begin{equation}
\mathcal{L}_{\text{lb}} = \lambda_{\text{lb}} \; \alpha_{\text{lb}}(t) \; \text{KL}\bigl(\bar{\mathbf{w}} \;\|\; \mathbf{u}\bigr)
\label{eq:loadbalance}
\end{equation}
where $\bar{\mathbf{w}}$ is the batch-and-group-averaged gating weight:
\begin{equation}
\bar{w}_k = \frac{1}{B} \sum_{b=1}^{B} \frac{1}{G} \sum_{g=1}^{G} w_{k,g}(\mathbf{s}_t^{(b)})
\label{eq:wbar_def}
\end{equation}
$\mathbf{u} = [1/K, \ldots, 1/K]$ is the uniform prior, $\lambda_{\text{lb}}$ is the coefficient, and $\alpha_{\text{lb}}(t)$ is an annealing factor that decays during training. This regularizer constrains only batch-level statistics, preventing permanent deactivation while allowing state-dependent specialization. The overall actor loss is:
\begin{equation}
\mathcal{L}_{\text{actor}} = \mathbb{E}_{\mathbf{s}_t \sim \mathcal{D}}\bigl[\alpha \log \pi_\theta(\mathbf{a}_t | \mathbf{s}_t) - Q_\phi(\mathbf{s}_t, \mathbf{a}_t)\bigr] + \mathcal{L}_{\text{lb}}
\label{eq:actorloss}
\end{equation}
where $\alpha$ is the SAC entropy coefficient, and $\theta$ and $\phi$ denote the learnable parameters of the actor (experts, recruitment network, $S_{k,j}$, and $c_k$) and of the critic, respectively. Here $Q_\phi$ is the scalar value induced by the distributional critic of Sec.~\ref{sec:critic}, i.e., the expectation $Q_\phi(\mathbf{s},\mathbf{a}) = \mathbb{E}[Z(\mathbf{s},\mathbf{a})] = \sum_i z_i \, p_i(\mathbf{s},\mathbf{a})$ of the aggregated distribution in Eq.~\eqref{eq:critic_agg}.

\subsection{Expert-Aware Distributional Critic}
\label{sec:critic}

A standard scalar critic is insufficient to capture the heterogeneous contributions of multiple experts to the expected return. We propose an expert-aware distributional critic with per-expert value heads that model return distributions rather than point estimates.

\textbf{Per-expert distributional heads.}
For each of $M=2$ critics and each expert $k$, we define a categorical distribution over atoms $\{z_1, \ldots, z_{N_{\text{atom}}}\}$ following C51~\cite{bellemare2017distributional}:
\begin{equation}
Z_{m,k}(\mathbf{s}, \mathbf{a}) = \sum_{i=1}^{N_{\text{atom}}} p_{m,k,i}(\mathbf{s}, \mathbf{a}) \; \delta(z_i)
\label{eq:dist_head}
\end{equation}
where $p_{m,k,i}(\mathbf{s}, \mathbf{a}) \in [0,1]$ are softmax-output probabilities for each atom.

\textbf{Gating-weighted aggregation.}
Per-expert distributions are aggregated using the actor's gating weights:
\begin{equation}
Z_m(\mathbf{s}, \mathbf{a}) = \sum_{k=1}^{K} \tilde{w}_k(\mathbf{s}) \; Z_{m,k}(\mathbf{s}, \mathbf{a})
\label{eq:critic_agg}
\end{equation}
where $\tilde{w}_k(\mathbf{s}) = \frac{1}{G} \sum_{g=1}^{G} w_{k,g}(\mathbf{s})$ is the per-state group-averaged gating weight (distinct from the batch-level statistic $\bar{w}_k$ in Eq.~\eqref{eq:wbar_def}), directly reusing the actor's gating weights --- no separate critic gating is learned. This makes the critic's value structure isomorphic to the actor's expert structure: the same gating that determines each expert's action contribution also determines its value contribution, enabling fine-grained credit assignment.

\textbf{Distributional Bellman training.}
The critic is trained via distributional Bellman projection with cross-entropy loss:
\begin{equation}
\mathcal{L}_{\text{critic}} = \sum_{m=1}^{M} \text{CE}\bigl(\Pi \, \mathcal{T}^\pi Z'_m, \; Z_m\bigr)
\label{eq:criticloss}
\end{equation}
where $Z'_m$ is the distribution predicted by a slowly updated target critic, the distributional Bellman operator is defined as $(\mathcal{T}^\pi Z)(\mathbf{s}_t, \mathbf{a}_t) \overset{D}{=} r_t + \gamma Z(\mathbf{s}_{t+1}, \mathbf{a}_{t+1})$ with $\mathbf{a}_{t+1} \sim \pi_\theta(\cdot \mid \mathbf{s}_{t+1})$, and $\Pi$ is the categorical projection that distributes the probability mass of each shifted atom $r_t + \gamma z_i$ onto its two nearest neighbors in the fixed support $\{z_1, \ldots, z_{N_{\text{atom}}}\}$ by linear interpolation~\cite{bellemare2017distributional}. This design provides three benefits: (1) distributional information reduces value estimation bias in high-dynamic motions; (2) the isomorphic actor-critic structure enables per-expert credit assignment; and (3) full dual-critic SAC compatibility preserves training stability.

\subsection{Quota-Routed Replay with Deferred Scheduling}
\label{sec:replay}

\begin{figure}[t]
\centering
\includegraphics[width=0.92\columnwidth]{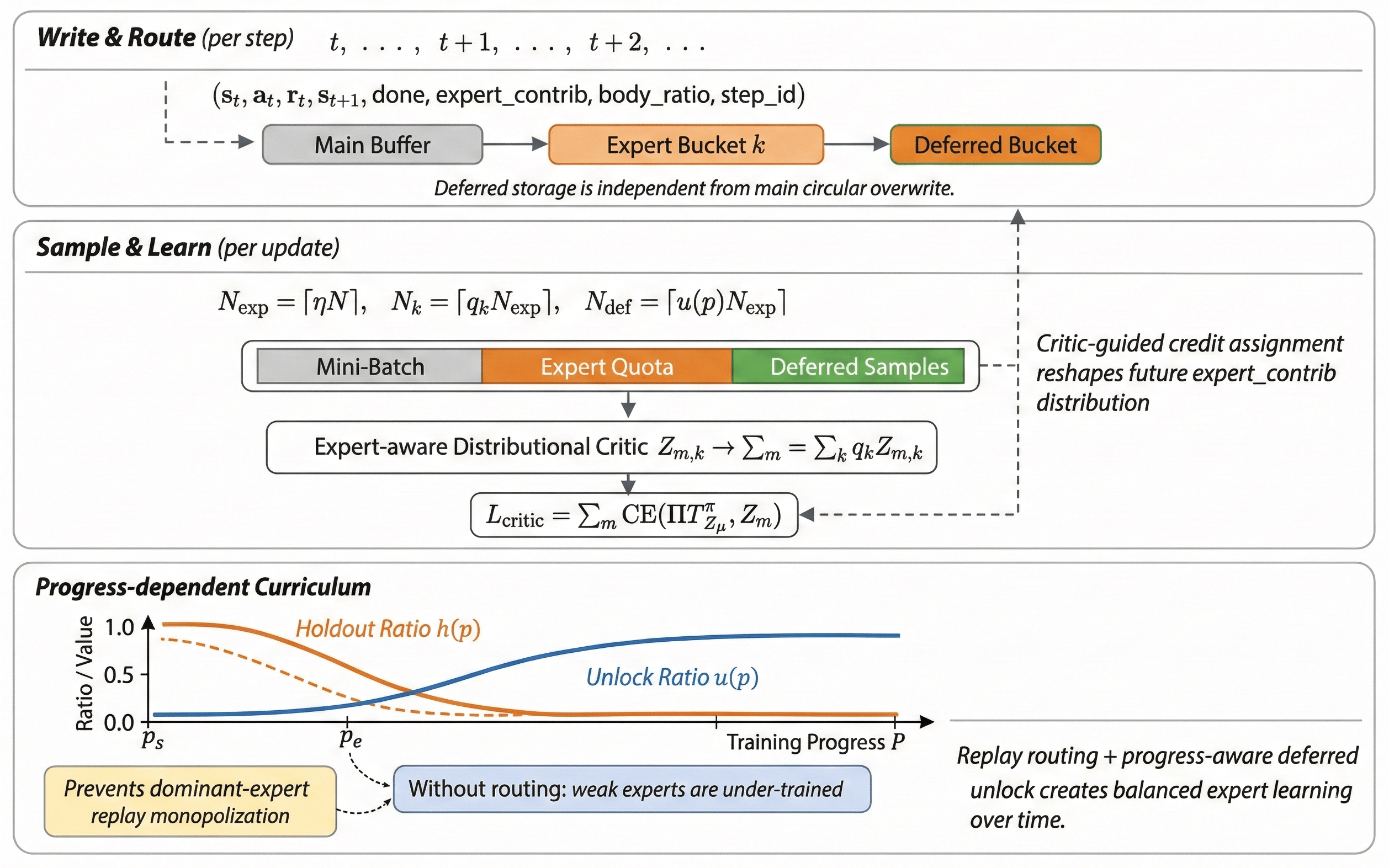}
\caption{\textbf{Quota-Routed Replay and Expert-Aware Critic over training time.} \emph{Top:} Each transition is stored with expert contribution metadata and routed to the main buffer, per-expert buckets, and deferred buckets. \emph{Middle:} Mini-batches combine uniform samples, expert-quota samples ($N_k = \lfloor q_k N_{\text{exp}} \rfloor$), and deferred samples ($N_{\text{def}} = \lfloor u(\rho) N_{\text{exp}} \rfloor$). \emph{Bottom:} Holdout ratio $h(\rho)$ releases held-back data early, while unlock ratio $u(\rho)$ ramps deferred samples in the middle and late phases.}
\label{fig:buffer}
\end{figure}

In multi-expert training, uniform replay causes transitions from dominant experts to be overrepresented, leaving weaker experts under-trained. We address this with quota-routed sampling combined with progress-dependent deferred scheduling (Fig.~\ref{fig:buffer}).

\textbf{Expert contribution tagging.}
Each stored transition is augmented with:
\begin{itemize}[leftmargin=*]
    \item \textbf{Expert contribution vector} $\mathbf{e}_t \in \mathbb{R}^K$, where $e_{t,k} = \frac{1}{d} \sum_{j} |w_{k,g(j)} c_k S_{k,j} \mu_{k,j}|$ measures the mean absolute action contribution of expert $k$.
    \item \textbf{Body tracking ratio} $b_t \in [0, 1]$, the fraction of key bodies below error threshold $\epsilon_{\text{body}}$.
    \item \textbf{Step identifier} $t_{\text{step}}$ for progress-dependent scheduling.
\end{itemize}
The dominant expert $k^* = \arg\max_k e_{t,k}$ determines routing; transitions with low $b_t$ are routed to deferred buckets.

\textbf{Quota-routed sampling.}
Let $N$ be the mini-batch size and $\eta \in [0,1]$ the expert-routing fraction:
\begin{equation}
N_{\text{exp}} = \lfloor \eta N \rfloor, \qquad N_k = \lfloor q_k \, N_{\text{exp}} \rfloor
\label{eq:quota}
\end{equation}
where $q_k$ is the target quota for expert $k$. This guarantees every expert receives training exposure proportional to its assigned responsibility, preventing the feedback loop where strong experts improve while weak experts stagnate.

\textbf{Deferred buckets.}
To prevent difficult samples from destabilizing early training, we maintain independent deferred buckets with separate capacity from the main buffer. Release is controlled by two progress-dependent schedules over $\rho \in [0, 1]$ (normalized training progress). The holdout ratio:
\begin{equation}
h(\rho) = h_{\max} \cdot \max\bigl(0, \; 1 - \rho / \rho_h\bigr)
\label{eq:holdout}
\end{equation}
The deferred unlock ratio:
\begin{equation}
u(\rho) = \begin{cases}
0 & \rho \leq \rho_s \\[4pt]
u_{\max} \cdot \dfrac{\rho - \rho_s}{\rho_e - \rho_s} & \rho_s < \rho < \rho_e \\[4pt]
u_{\max} & \rho \geq \rho_e
\end{cases}
\label{eq:unlock}
\end{equation}
yielding $N_{\text{def}} = \lfloor u(\rho) \cdot N_{\text{exp}} \rfloor$ deferred samples per batch. This transforms the decision of when to train on difficult samples from a static hyperparameter into a dynamic, progress-aware strategy: stable convergence in the early phase, with deferred samples gradually released to raise the performance ceiling.

\subsection{Closed-Loop Training}
\label{sec:loop}

The three mechanisms are not independent modules but form an end-to-end closed training loop (Fig.~\ref{fig:framework}):
\[
\text{Actor} \xrightarrow{\texttt{contrib}} \text{Routing} \xrightarrow{\text{quota}} \text{Critic} \xrightarrow{\text{grad}} \text{Actor}
\]
The MoE actor generates actions with per-expert contribution profiles $\mathbf{e}_t$, which determine replay routing. The quota sampler draws balanced mini-batches for the distributional critic, whose gradients refine expert specialization and future contribution profiles. This closed loop ensures that policy decomposition, data allocation, and value learning are mutually reinforcing rather than operating in isolation.

LooperMuscle extends the FastSAC recipe along three orthogonal axes --- policy structure, value estimation, and data allocation --- coupled into a single closed training loop. All hyperparameters are provided in our released codebase.
\section{Experiments}
\label{sec:experiments}

We evaluate \textbf{LooperMuscle} on 29-DoF humanoid whole-body tracking and focus on the speed--performance tradeoff against PPO~\cite{schulman2017proximal} and FastSAC~\cite{seo2025learning} baselines. We answer the following questions:
\begin{enumerate}
    \item[\textbf{Q1}] Can LooperMuscle substantially improve FastSAC-level tracking quality while keeping a large wall-clock advantage over PPO?
    \item[\textbf{Q2}] Are the gains consistent across motion categories?
    \item[\textbf{Q3}] Do the proposed components each contribute?
    \item[\textbf{Q4}] Does expert specialization emerge in real training curves?
\end{enumerate}

\subsection{Experimental Setup}
\label{sec:exp_setup}

\textbf{Simulation platform.}
All quantitative benchmarking is performed in MJLab~\cite{zakka2026mjlab} on the Unitree G1 humanoid (29~DoF), with a simulation timestep of 0.005\,s, control decimation of 4 (yielding a 50\,Hz control frequency), and 10\,s episodes (500 control steps per episode). Training uses 4096 parallel environments on a single NVIDIA RTX 4090D.

\textbf{Dataset.}
We benchmark on 40 LAFAN1~\cite{harvey2020robust} sequences covering six categories: Walk (12), Run/Sprint (6), Jump (3), Dance (8), Fight (5), and Fall\&GetUp (6).

\textbf{Compared methods.}
We compare:
\begin{itemize}[leftmargin=*]
    \item \textbf{PPO}~\cite{schulman2017proximal} --- on-policy reference, trained for $\sim$6\,h;
    \item \textbf{FastSAC-MLP}~\cite{seo2025learning} --- off-policy FastSAC with a monolithic MLP actor, trained for $\sim$15\,min;
    \item \textbf{LooperMuscle} --- FastSAC backbone augmented with MoE actor ($K{=}4$ experts), expert-aware distributional critic, and quota-routed replay with deferred scheduling, trained for $\sim$45\,min.
\end{itemize}
PPO serves as the quality upper bound. FastSAC-MLP and LooperMuscle are the competing fast-training methods.

\textbf{Metrics.}
We report: (i)~Mean Body Position Error (m, lower is better), which measures the average Euclidean distance between tracked key-body positions and their reference targets; (ii)~Mean Joint Position Error (rad, lower is better), which measures the average absolute deviation of joint angles from their reference values; (iii)~Normalized Reward (PPO converged score normalized to 1.0); and (iv)~wall-clock training time. Reported errors are means over the sequences of each category (std across sequences in Table~\ref{tab:main_results}; per-category std as error bars in Fig.~\ref{fig:per_category_body}). All runs are repeated over multiple random seeds; shaded regions and ellipses in Figs.~\ref{fig:teaser} and \ref{fig:main_results} denote $\pm 1$ std across seeds.

\textbf{Configuration note.}
LooperMuscle uses $K=4$ experts, two-group gating ($G{=}2$, i.e., upper-body and lower-body), KL-divergence-based gate load-balance regularization (Eq.~\eqref{eq:loadbalance}), contribution-based replay routing (Sec.~\ref{sec:replay}), and deferred unlock scheduling (Eq.~\eqref{eq:unlock}). A single fixed hyperparameter setting ($\tau_g$, $\lambda_{\text{lb}}$ and its annealing schedule, routing fraction $\eta$, and curriculum thresholds) is used for all 40 sequences without per-motion tuning; in our experience these values transfer across motion categories without delicate re-tuning. Full values are provided in the released codebase.

\textbf{Integrity disclosure on observation interfaces.}
Main quantitative tables and plots are from MJLab~\cite{zakka2026mjlab} benchmarking, whose observation interface includes ground-truth anchor-point perception (global key-body positions and orientations queried from the simulator state) that cannot be measured on real hardware. Real-robot deployment instead uses the open-source Holosoma stack~\cite{amazonfar2025holosoma} with a 154-dimensional interface built only from onboard-estimable signals; the hardware policy is therefore retrained from scratch under this interface (Sec.~\ref{sec:hardware_deploy}). MJLab numbers should thus be read as controlled algorithmic comparisons under a shared privileged interface rather than direct predictors of hardware performance; hardware results are reported as \textbf{deployment validation} and are not pooled into MJLab benchmark rankings.

\subsection{Main Results: Speed--Quality Tradeoff}
\label{sec:main_results}
\begin{figure*}[t]
\centering
\begin{subfigure}[t]{0.25\textwidth}
    \centering
    \includegraphics[width=\textwidth]{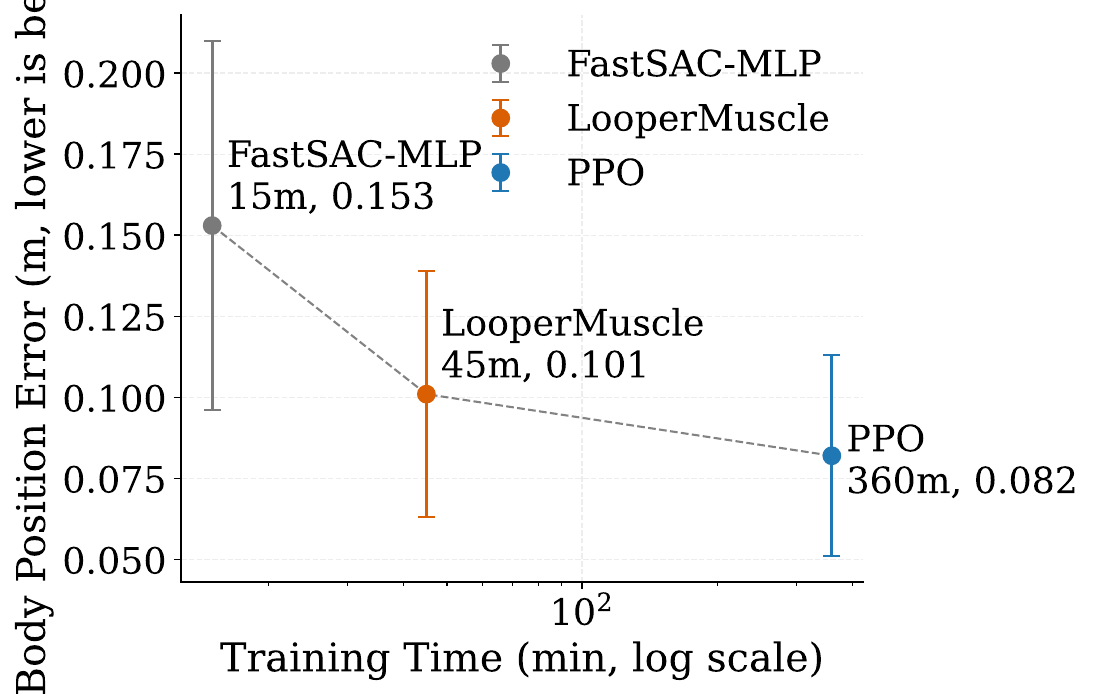}
    \caption{Speed--quality tradeoff.}
    \label{fig:speed_quality_tradeoff}
\end{subfigure}
\hfill
\begin{subfigure}[t]{0.25\textwidth}
    \centering
    \includegraphics[width=\textwidth]{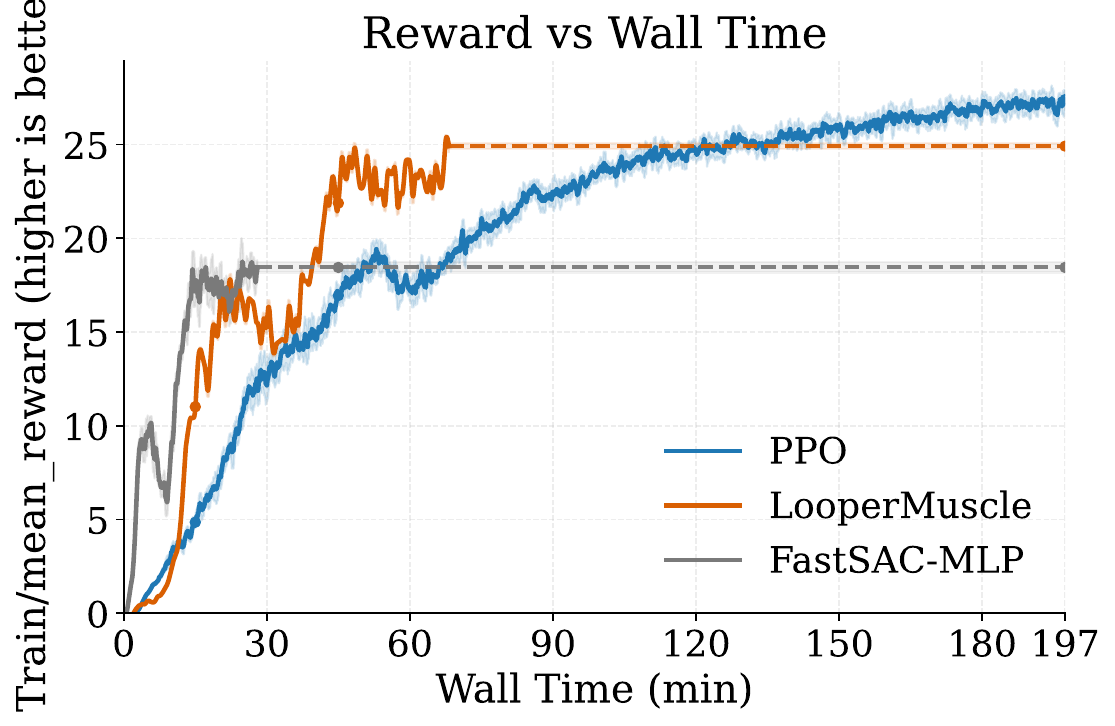}
    \caption{Reward vs wall-clock time.}
    \label{fig:checkpoint_curve}
\end{subfigure}
\hfill
\begin{subfigure}[t]{0.24\textwidth}
    \centering
    \includegraphics[width=\textwidth]{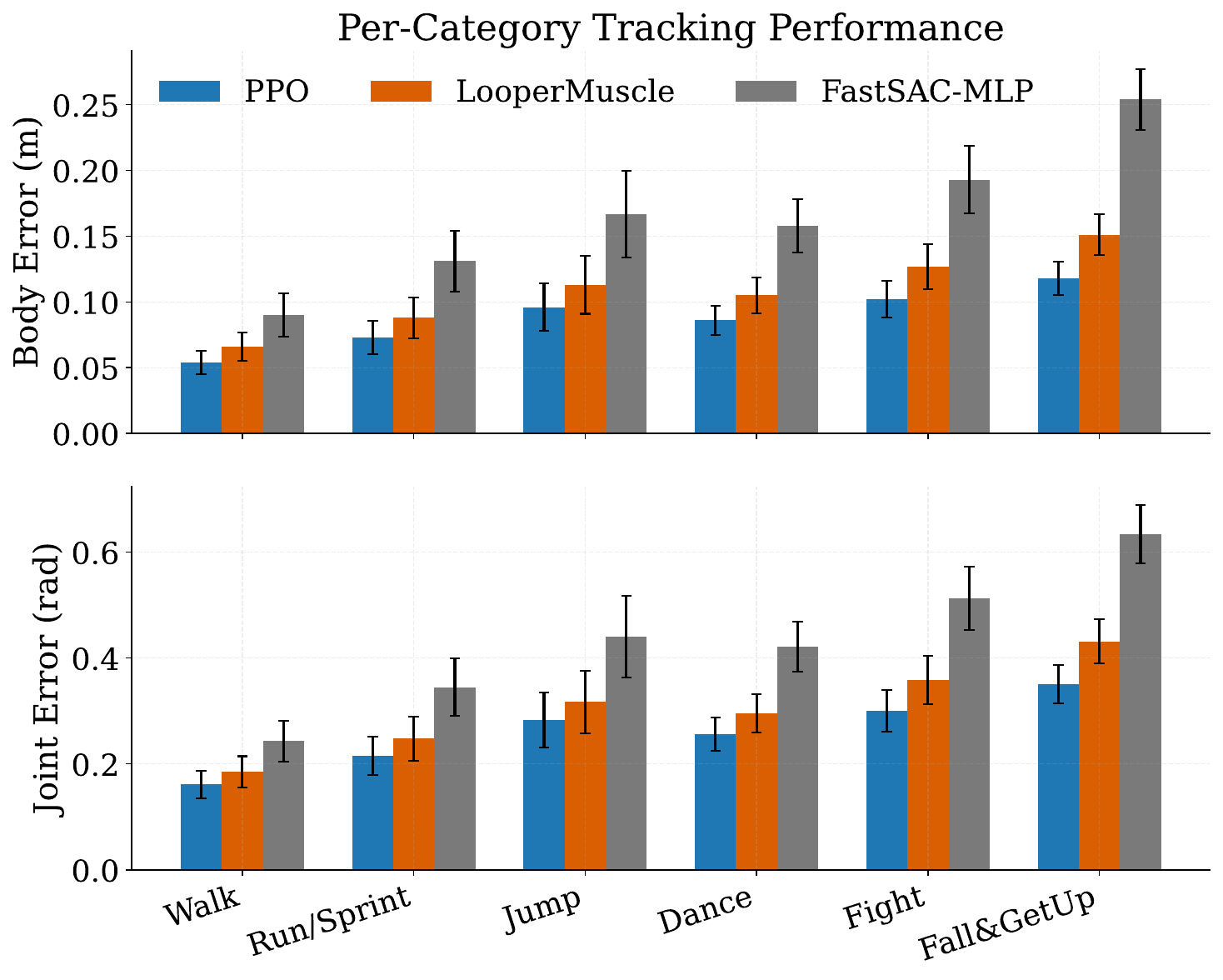}
    \caption{Per-category body/joint error.}
    \label{fig:per_category_body}
\end{subfigure}
\hfill
\begin{subfigure}[t]{0.22\textwidth}
    \centering
    \includegraphics[width=\textwidth]{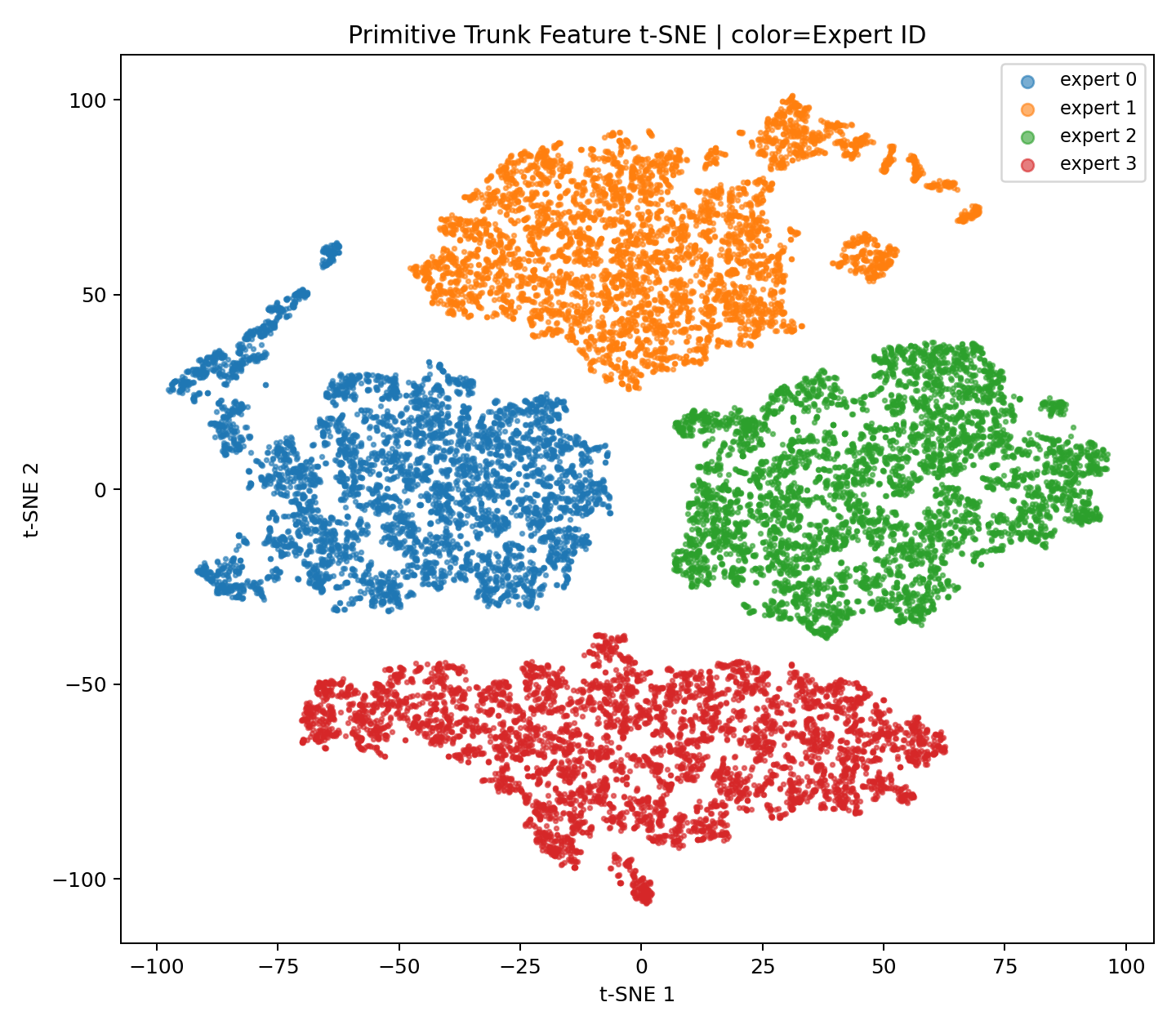}
    \caption{Expert specialization.}
    \label{fig:expert_tsne}
\end{subfigure}
\caption{\textbf{Main quantitative results and expert specialization.} (a)~LooperMuscle occupies a favorable middle point in the speed--quality tradeoff: much better tracking quality than FastSAC-MLP and much less training time than PPO. Ellipses denote $\pm 1$ standard deviation over multiple random seeds. (b)~Reward vs.\ wall-clock time (0--197\,min) under the same reference task in the short-budget regime. (c)~Per-category body/joint error with $\pm 1$ standard deviation bars; LooperMuscle consistently improves over FastSAC-MLP, with larger gains on high-difficulty categories. (d)~t-SNE~\cite{van2008visualizing} projection of primitive trunk features (for each state, the penultimate-layer activation of its dominant expert), colored by dominant expert ID; states routed to different experts occupy well-separated regions, with clusters aligning to dynamic regimes (e.g., stance vs.\ swing) rather than motion categories.}
\label{fig:main_results}
\end{figure*}

\begin{table}[h]
\centering
\caption{Overall comparison on 40 LAFAN1 sequences. \textbf{Bold} denotes best among fast-training methods ($\leq$1\,h). PPO is shown as the quality reference only.}
\label{tab:main_results}
\resizebox{0.48\textwidth}{!}{
\begin{tabular}{lcccc}
\toprule
Method & Body Err. (m)$\downarrow$ & Joint Err. (rad)$\downarrow$ & Norm. Reward$\uparrow$ & Time \\
\midrule
\multicolumn{5}{l}{\textit{Reference ($\sim$6\,h training):}} \\
PPO           & 0.082$\pm$0.031 & 0.243$\pm$0.089 & 1.000 & $\sim$360 min \\
\midrule
\multicolumn{5}{l}{\textit{Fast methods ($\leq$1\,h training):}} \\
LooperMuscle  & \textbf{0.101$\pm$0.038} & \textbf{0.285$\pm$0.102} & \textbf{0.723} & $\sim$45 min \\
FastSAC-MLP   & 0.153$\pm$0.057 & 0.401$\pm$0.134 & 0.648 & $\sim$15 min \\
\bottomrule
\end{tabular}
}
\end{table}

As shown in Table~\ref{tab:main_results}, LooperMuscle reduces body-position tracking error by 34.0\% relative to FastSAC-MLP (0.153$\rightarrow$0.101\,m), while remaining much faster than PPO in wall-clock training time.

\begin{figure*}[t]
\centering
\includegraphics[width=0.92\textwidth]{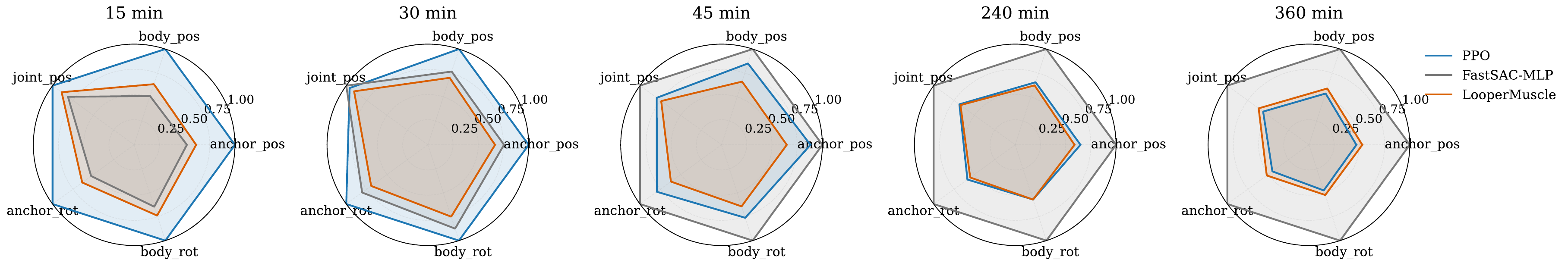}
\caption{\textbf{Tracking error radar profiles at 15\,min, 30\,min, 45\,min, 240\,min, and 360\,min} (lower area is better). This time-profiled view shows how method-level error structure evolves from early to late training under the same task setup.}
\label{fig:tracking_radar_time}
\end{figure*}

\textbf{Effect of additional training budget.}
Since LooperMuscle trains for $\sim$45\,min versus FastSAC-MLP's $\sim$15\,min, a natural question is whether the baseline would catch up given an equal budget. Our results indicate it does not: all methods in Fig.~\ref{fig:teaser} (top) are trained for the same number of steps, and Figs.~\ref{fig:checkpoint_curve} and \ref{fig:tracking_radar_time} evaluate FastSAC-MLP up to 197 and 360\,min of wall-clock training; its reward saturates shortly after the 15-minute mark and subsequent gains are marginal, leaving the gap essentially unchanged. This points to a representational rather than compute bottleneck in the monolithic actor, consistent with Sec.~\ref{sec:ablation}; a parameter-matched larger MLP baseline is a valuable further control left for future work.

\textbf{Relation to locomotion results in \cite{seo2025learning}.}
FastSAC was reported to match PPO on locomotion, whereas PPO remains the stronger reference here. We attribute this to the task: locomotion tracks low-dimensional velocity commands, while WBT must track dense, time-varying references over all 29 joints and key bodies --- a far harder credit-assignment problem in which massively parallel on-policy PPO retains an advantage.

\begin{figure*}[t]
\centering
\includegraphics[width=0.92\textwidth]{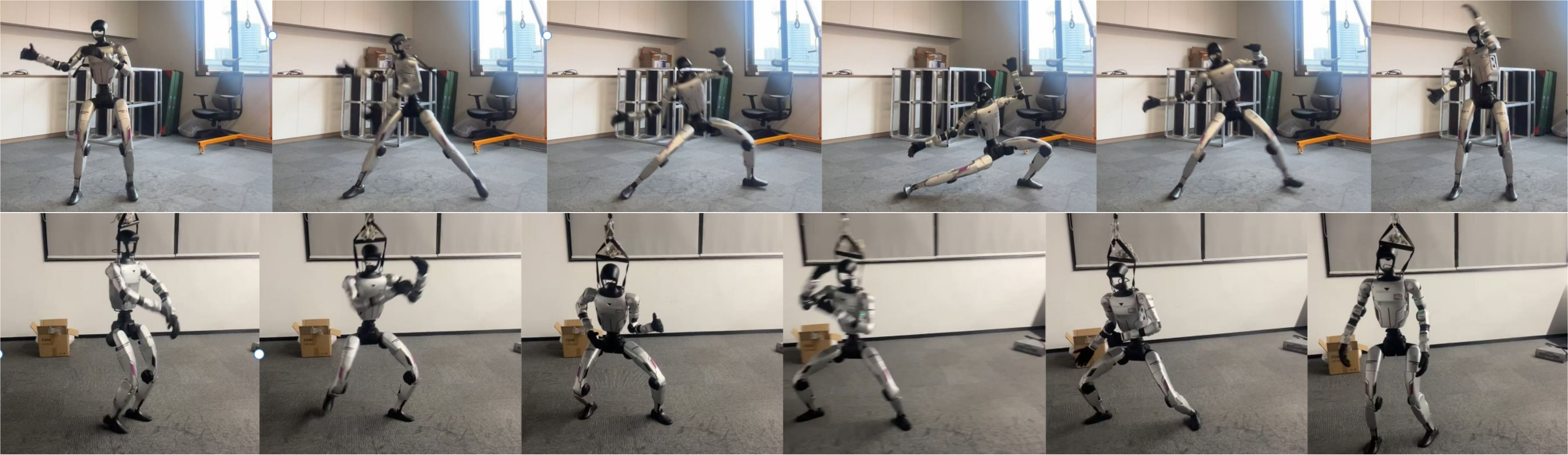}
\caption{\textbf{Real-world deployment on Unitree G1.} The LooperMuscle policy tracks the whole-body motion of a fighting sequence from the KungfuBot2 motion library~\cite{han2025kungfubot2}, with stable balance and smooth transitions between kicks, demonstrating sim-to-real transfer via the Holosoma runtime stack at 50\,Hz.}
\label{fig:real_deploy}
\end{figure*}

\subsection{Per-Category Results}
\label{sec:per_category}

\begin{table}[t]
\centering
\caption{Per-category performance on 40 sequences. \textbf{Bold} denotes best among fast-training methods. PPO is quality reference only.}
\label{tab:per_category}
\resizebox{\columnwidth}{!}{
\begin{tabular}{l c cc cc cc c}
\toprule
& & \multicolumn{2}{c}{PPO} & \multicolumn{2}{c}{LooperMuscle} & \multicolumn{2}{c}{FastSAC-MLP} & Ours vs FastSAC \\
\cmidrule(lr){3-4} \cmidrule(lr){5-6} \cmidrule(lr){7-8}
Category & \#Seq & Body$\downarrow$ & Joint$\downarrow$ & Body$\downarrow$ & Joint$\downarrow$ & Body$\downarrow$ & Joint$\downarrow$ & Body Impr. \\
\midrule
Walk        & 12 & 0.054 & 0.161 & \textbf{0.066} & \textbf{0.185} & 0.090 & 0.243 & $\downarrow$26.7\% \\
Run/Sprint  &  6 & 0.073 & 0.215 & \textbf{0.088} & \textbf{0.248} & 0.131 & 0.345 & $\downarrow$32.8\% \\
Jump        &  3 & 0.096 & 0.283 & \textbf{0.113} & \textbf{0.317} & 0.167 & 0.440 & $\downarrow$32.3\% \\
Dance       &  8 & 0.086 & 0.256 & \textbf{0.105} & \textbf{0.296} & 0.158 & 0.421 & $\downarrow$33.5\% \\
Fight       &  5 & 0.102 & 0.300 & \textbf{0.127} & \textbf{0.358} & 0.193 & 0.513 & $\downarrow$34.2\% \\
Fall\&GetUp &  6 & 0.118 & 0.350 & \textbf{0.151} & \textbf{0.431} & 0.254 & 0.634 & $\downarrow$40.6\% \\
\midrule
\textbf{All} & 40 & 0.082 & 0.243 & \textbf{0.101} & \textbf{0.285} & 0.153 & 0.401 & $\downarrow$34.0\% \\
\bottomrule
\end{tabular}
}
\end{table}

\textbf{Category-wise trend.}
Table~\ref{tab:per_category} and Fig.~\ref{fig:per_category_body} show that LooperMuscle consistently outperforms FastSAC-MLP across all six motion categories. The body-error reduction ranges from 26.7\% (Walk) to 40.6\% (Fall\&GetUp), with an overall gain of 34.0\% on all 40 sequences.

\textbf{Difficulty-dependent gain.}
The improvement broadly increases with motion difficulty: the margin is smallest on Walk (26.7\%) and largest on highly dynamic, contact-rich Fall\&GetUp (40.6\%). This trend indicates that the proposed expert specialization and replay curriculum are especially effective when control objectives are strongly coupled and non-stationary.

\subsection{Ablation Study}
\label{sec:ablation}

\begin{table}[h]
\centering
\caption{Ablation over 40 sequences. Each row removes a single component from the full system.}
\label{tab:ablation}
\resizebox{0.48\textwidth}{!}{
\begin{tabular}{lccc}
\toprule
Configuration & Body Err. (m)$\downarrow$ & Joint Err. (rad)$\downarrow$ & $\Delta$Body \\
\midrule
LooperMuscle (full)     & 0.101 & 0.285 & --- \\
\midrule
w/o MoE Actor           & 0.153 & 0.401 & +51.5\% \\
w/o Quota Replay        & 0.127 & 0.358 & +25.7\% \\
w/o Deferred Scheduling & 0.119 & 0.341 & +17.8\% \\
w/o Expert-Aware Critic & 0.113 & 0.324 & +11.9\% \\
\bottomrule
\end{tabular}
}
\end{table}

\textbf{Overall impact.}
Table~\ref{tab:ablation} confirms that all proposed modules are necessary. Removing the MoE Actor causes the largest degradation (+51.5\% body error), indicating that expert decomposition is the primary source of representational gain in high-DoF WBT. Note that removing the MoE actor also removes the expert structure on which quota routing and the expert-aware critic are defined, so this configuration reduces to the FastSAC-MLP baseline (hence the identical values in Table~\ref{tab:main_results}).

\textbf{Replay-side contributions.}
Removing Quota Replay increases body error by 25.7\%, showing that balanced expert exposure is critical for avoiding expert under-training. Independently removing Deferred Scheduling degrades performance by 17.8\%, suggesting that progressive sample release improves optimization stability, especially for hard-motion transitions.

\textbf{Value-learning contribution.}
Replacing the expert-aware critic with a standard (non-expert) distributional critic still causes a notable drop (+11.9\%), which supports the claim that value estimation aligned with expert structure improves credit assignment.

\textbf{Non-additive effect.}
The degradations are not strictly additive, indicating coupling among modules: actor specialization, routed replay, and expert-aware value learning form a closed-loop training system rather than independent heuristics.

\subsection{Training Curves and Expert Specialization}
\label{sec:training_curves_expert}

Fig.~\ref{fig:teaser} (top) plots the training reward curves for PPO, FastSAC-MLP, and LooperMuscle on a matched reference motion, providing a direct comparison of convergence behavior.

To verify that expert decomposition produces meaningful specialization, we visualize the \emph{primitive trunk features} --- for each state, the penultimate-layer (trunk) activation of its dominant expert $k^* = \arg\max_k e_{t,k}$ --- via t-SNE~\cite{van2008visualizing} (Fig.~\ref{fig:expert_tsne}). States routed to different dominant experts occupy well-separated regions, confirming that the anti-collapse regularizer drives non-redundant specialization. Notably, clusters align with dynamic regimes (e.g., stance vs.\ swing phase) rather than motion categories, indicating that experts specialize by control mode.

Fig.~\ref{fig:tracking_radar_time} reports radar profiles at five checkpoints. At 15\,min, FastSAC-MLP has already converged; by 45\,min LooperMuscle surpasses it on most axes, converting additional training cost into substantial quality gain. At 240 and 360\,min PPO reaches the best final profile, while LooperMuscle remains close with much lower wall-clock cost. The temporal evolution aligns with the ablation findings: early stability is aided by replay scheduling, mid-stage gains by expert specialization, and late-stage robustness by expert-aware value learning.

\subsection{Comparison with Concurrent MoE-Based Methods}
\label{sec:comparison_concurrent}

We contextualize LooperMuscle by examining whether the benefit of MoE over monolithic MLP is consistent across different training paradigms. Table~\ref{tab:cross_validation} summarizes the MLP-to-MoE relative gains reported in concurrent work and in our own experiments.

\begin{table}[t]
\centering
\caption{MoE consistently outperforms monolithic MLP across different training regimes. Each regime row uses a \emph{different} dataset, DoF, and metric; therefore, \textbf{only the relative improvement ($\Delta$ vs MLP) within each regime is meaningful} --- absolute tracking errors are \textbf{not comparable} across regimes.}
\label{tab:cross_validation}
\resizebox{0.48\textwidth}{!}{
\begin{tabular}{llcl}
\toprule
Regime & Configuration & $\Delta$ vs MLP Baseline & Source \\
\midrule
\multicolumn{4}{l}{\textit{Regime A: PPO, AMASS 9770 seq, 23-DoF, $E_{\text{mpkpe}}$ (mm$\downarrow$)}} \\
PPO     & MLP (ExBody2) --- \emph{baseline}   & ---              & \cite{ji2024exbody2} \\
PPO     & Motion MoE (GMT)           & $\downarrow$14.6\% & \cite{chen2025gmt} \\
PPO     & MLP (VMS-MLP)              & $\downarrow$7.4\%  & \cite{han2025kungfubot2} \\
PPO     & Soft MoE (VMS-MoE)         & $\downarrow$15.1\% & \cite{han2025kungfubot2} \\
PPO     & Orth.\ MoE (\textbf{VMS/KungfuBot2}) & $\downarrow$19.7\% & \cite{han2025kungfubot2} \\
\midrule
\multicolumn{4}{l}{\textit{Regime B: FastSAC, LAFAN1 40 seq, 29-DoF, Body Pos.\ Err.\ (m$\downarrow$)}} \\
FastSAC & MLP (FastSAC-MLP) --- \emph{baseline}   & ---              & Ours \\
FastSAC & Group MoE (\textbf{LooperMuscle})   & $\downarrow$34.0\% & Ours \\
\bottomrule
\end{tabular}
}
\end{table}

\textbf{Regimes.}
In Regime A, ExBody2~\cite{ji2024exbody2}, GMT~\cite{chen2025gmt}, and VMS (KungfuBot2)~\cite{han2025kungfubot2} are concurrent PPO-based methods trained on the large-scale AMASS dataset (9,770 sequences, 23-DoF, $E_{\text{mpkpe}}$ in mm); ExBody2 uses a monolithic MLP actor and serves as the regime's baseline; GMT combines a motion MoE with adaptive motion sampling, and VMS introduces Orthogonal MoE ($K{=}6$) with Gram--Schmidt orthogonalization, reducing tracking error by 14.6\% and 19.7\%, respectively, whereas VMS's own MLP variant improves by only 7.4\%. In Regime B --- our fundamentally different off-policy FastSAC setting (LAFAN1, 40 sequences, 29-DoF, body position error in m) --- our Group MoE reduces tracking error by 34.0\% over FastSAC-MLP.

\textbf{What can and cannot be concluded.}
The two regimes differ in training algorithm (PPO vs.\ FastSAC), dataset scale (9,770 vs.\ 40 sequences), robot DoF (23 vs.\ 29), and evaluation metric, precluding direct comparison of absolute errors or gain magnitudes. However, the consistent \emph{direction} of improvement --- MoE substantially outperforming monolithic MLPs in both on-policy and off-policy paradigms --- suggests that expert decomposition is a broadly beneficial design principle for humanoid whole-body tracking rather than an artifact of a particular setup.

\subsection{Hardware Deployment Validation}
\label{sec:hardware_deploy}

We validate LooperMuscle on a physical Unitree G1 using the open-source Holosoma runtime stack~\cite{amazonfar2025holosoma}. The key difference from the MJLab benchmark is the observation interface: MJLab provides ground-truth anchor-point perception (global key-body positions and orientations queried directly from the simulator state) that no onboard sensor can measure. The Holosoma interface removes these privileged signals and retains only onboard-estimable quantities (e.g., IMU base orientation and angular velocity, joint encoder readings, and motion-reference features expressed relative to the robot base), resulting in a 154-dimensional observation vector. Due to this interface difference, the hardware policy is \textbf{retrained from scratch} in Holosoma --- using the same LooperMuscle architecture, components, and training recipe, with only the observation interface changed --- rather than directly transferring the MJLab checkpoint. The retrained policy outputs target joint positions $\mathbf{a}_t \in \mathbb{R}^{29}$ at 50\,Hz via ONNX inference. As shown in Fig.~\ref{fig:teaser} (bottom) and Fig.~\ref{fig:real_deploy}, the deployed policy executes whole-body Kungfu sequences from the KungfuBot2 motion library~\cite{han2025kungfubot2} with stable balance and smooth transitions. This validates that the LooperMuscle recipe trains deployable policies under a realistic observation interface; repeating the full benchmark and ablation suite under this deployable interface is an important next step (Sec.~\ref{sec:conclusion}).
\section{Conclusion}
\label{sec:conclusion}

LooperMuscle couples a per-joint-group MoE actor, per-expert distributional critic, and quota-routed replay into a closed training loop that narrows the speed--performance gap in humanoid whole-body tracking. It substantially improves tracking quality over FastSAC while remaining much faster than PPO, and deployment on a physical Unitree G1 validates real-world applicability. Several limitations remain. First, our quantitative benchmark and ablations use MJLab's privileged observation interface, whereas the hardware policy is retrained under a deployable interface; repeating the benchmark and ablation suite under the deployable interface, and quantifying the associated retraining cost, is our most immediate next step toward a unified sim-to-real pipeline. Second, parameter-matched baselines would further isolate the contribution of expert structure from capacity. Third, several design choices (expert count $K$, group partition $G$, per-group temperatures, routing fraction) are fixed rather than systematically varied, and evaluation covers a single robot morphology. Future work will additionally explore adaptive expert allocation, systematic domain randomization, and diffusion-based motion priors.
\bibliographystyle{IEEEtran}
\bibliography{ref}

@inproceedings{rudin2022learning,
  title={Learning to Walk in Minutes Using Massively Parallel Deep Reinforcement Learning},
  author={Rudin, Nikita and Hoeller, David and Reist, Philipp and Hutter, Marco},
  booktitle={Proceedings of the 5th Conference on Robot Learning},
  volume={164},
  pages={91--100},
  year={2022},
  publisher={PMLR}
}

@article{makoviychuk2021isaac,
  title={Isaac Gym: High Performance GPU-Based Physics Simulation for Robot Learning},
  author={Makoviychuk, Viktor and Wawrzyniak, Lukasz and Guo, Yunrong and Lu, Michelle and Storey, Kier and Macklin, Miles and Hoeller, David and Rudin, Nikita and Allshire, Arthur and Handa, Ankur and State, Gavriel},
  year={2021}
}

@article{zakka2025mujoco,
  title={MuJoCo Playground},
  author={Zakka, Kevin and Tabanpour, Baruch and Liao, Qiayuan and Haiderbhai, Mustafa and Holt, Samuel and Luo, Jing Yuan and Allshire, Arthur and Frey, Erik and Sreenath, Koushil and Kahrs, Lueder A and others},
  journal={arXiv preprint arXiv:2502.08844},
  year={2025}
}

@inproceedings{zhao2020sim,
  title={Sim-to-Real Transfer in Deep Reinforcement Learning for Robotics: A Survey},
  author={Zhao, Wenshuai and Queralta, Jorge Pe{\~n}a and Westerlund, Tomi},
  booktitle={2020 IEEE Symposium Series on Computational Intelligence (SSCI)},
  year={2020}
}

@inproceedings{chebotar2019closing,
  title={Closing the Sim-to-Real Loop: Adapting Simulation Randomization with Real World Experience},
  author={Chebotar, Yevgen and Handa, Ankur and Makoviychuk, Viktor and Macklin, Miles and Issac, Jan and Ratliff, Nathan and Fox, Dieter},
  booktitle={2019 International Conference on Robotics and Automation (ICRA)},
  year={2019}
}

@article{seo2025fasttd3,
  title={FastTD3: Simple, Fast, and Capable Reinforcement Learning for Humanoid Control},
  author={Seo, Younggyo and Sferrazza, Carmelo and Geng, Haoran and Nauman, Michal and Yin, Zhao-Heng and Abbeel, Pieter},
  journal={arXiv preprint arXiv:2505.22642},
  year={2025}
}

@article{seo2025learning,
  title={Learning Sim-to-Real Humanoid Locomotion in 15 Minutes},
  author={Seo, Younggyo and Sferrazza, Carmelo and Chen, Juyue and Shi, Guanya and Duan, Rocky and Abbeel, Pieter},
  journal={arXiv preprint arXiv:2512.01996},
  year={2025}
}

@article{schulman2017proximal,
  title={Proximal Policy Optimization Algorithms},
  author={Schulman, John and Wolski, Filip and Dhariwal, Prafulla and Radford, Alec and Klimov, Oleg},
  journal={arXiv preprint arXiv:1707.06347},
  year={2017}
}

@inproceedings{bellemare2017distributional,
  title={A Distributional Perspective on Reinforcement Learning},
  author={Bellemare, Marc G and Dabney, Will and Munos, R{\'e}mi},
  booktitle={International Conference on Machine Learning},
  year={2017}
}

@article{peng2018deepmimic,
  title={DeepMimic: Example-Guided Deep Reinforcement Learning of Physics-Based Character Skills},
  author={Peng, Xue Bin and Abbeel, Pieter and Levine, Sergey and Van de Panne, Michiel},
  journal={ACM Transactions on Graphics (TOG)},
  year={2018}
}

@article{liao2025beyondmimic,
  title={BeyondMimic: From Motion Tracking to Versatile Humanoid Control via Guided Diffusion},
  author={Liao, Qiayuan and Truong, Takara E and Huang, Xiaoyu and Tevet, Guy and Sreenath, Koushil and Liu, C Karen},
  journal={arXiv preprint arXiv:2508.08241},
  year={2025}
}

@inproceedings{he2025hover,
  title={HOVER: Versatile Neural Whole-Body Controller for Humanoid Robots},
  author={He, Tairan and Xiao, Wenli and Lin, Toru and Luo, Zhengyi and Xu, Zhenjia and Jiang, Zhenyu and Kautz, Jan and Liu, Changliu and Shi, Guanya and Wang, Xiaolong and others},
  booktitle={2025 IEEE International Conference on Robotics and Automation (ICRA)},
  year={2025}
}

@inproceedings{he2025asap,
  title={ASAP: Aligning Simulation and Real-World Physics for Learning Agile Humanoid Whole-Body Skills},
  author={He, Tairan and Gao, Jiawei and Xiao, Wenli and Zhang, Yuanhang and Wang, Zi and Wang, Jiashun and Luo, Zhengyi and He, Guanqi and Sobanbab, Nikhil and Pan, Chaoyi and others},
  booktitle={Robotics: Science and Systems},
  year={2025}
}

@inproceedings{li2023parallel,
  title={Parallel Q-Learning: Scaling Off-Policy Reinforcement Learning under Massively Parallel Simulation},
  author={Li, Zechu and Chen, Tao and Hong, Zhang-Wei and Ajay, Anurag and Agrawal, Pulkit},
  booktitle={International Conference on Machine Learning},
  pages={19440--19459},
  year={2023},
  publisher={PMLR}
}

@misc{raffin2025sac,
  title={Getting SAC to Work on a Massive Parallel Simulator: An RL Journey with Off-Policy Algorithms},
  author={Raffin, Antonin},
  howpublished={araffin.github.io},
  year={2025}
}

@misc{shukla2025sac,
  title={Speeding Up SAC with Massively Parallel Simulation},
  author={Shukla, Arth},
  howpublished={arthshukla.substack.com},
  year={2025}
}

@inproceedings{dabney2018distributional,
  title={Distributional Reinforcement Learning with Quantile Regression},
  author={Dabney, Will and Rowland, Mark and Bellemare, Marc and Munos, R{\'e}mi},
  booktitle={Proceedings of the AAAI Conference on Artificial Intelligence},
  year={2018}
}

@inproceedings{peng2021amp,
  title={AMP: Adversarial Motion Priors for Stylized Physics-Based Character Control},
  author={Peng, Xue Bin and Ma, Ze and Abbeel, Pieter and Levine, Sergey and Kanazawa, Angjoo},
  booktitle={ACM SIGGRAPH},
  year={2021}
}

@article{luo2023perpetual,
  title={Perpetual Humanoid Control for Real-Time Simulated Avatars},
  author={Luo, Zhengyi and Cao, Jinkun and Weng, Alexander and Kitani, Kris and Xu, Weipeng},
  journal={arXiv preprint arXiv:2305.06456},
  year={2023}
}

@inproceedings{luo2024universal,
  title={Universal Humanoid Motion Representations for Physics-Based Control},
  author={Luo, Zhengyi and Cao, Jinkun and Merel, Josh and Winkler, Alexander and Huang, Jing and Kitani, Kris and Xu, Weipeng},
  booktitle={International Conference on Learning Representations (ICLR)},
  year={2024}
}

@article{shazeer2017outrageously,
  title={Outrageously Large Neural Networks: The Sparsely-Gated Mixture-of-Experts Layer},
  author={Shazeer, Noam and Mirhoseini, Azalia and Maziarz, Krzysztof and Davis, Andy and Le, Quoc and Hinton, Geoffrey and Dean, Jeff},
  journal={arXiv preprint arXiv:1701.06538},
  year={2017}
}

@article{fedus2022switch,
  title={Switch Transformers: Scaling to Trillion Parameter Models with Simple and Efficient Sparsity},
  author={Fedus, William and Zoph, Barret and Shazeer, Noam},
  journal={Journal of Machine Learning Research},
  volume={23},
  number={120},
  pages={1--39},
  year={2022}
}

@inproceedings{yang2020multi,
  title={Multi-Task Reinforcement Learning with Soft Modularization},
  author={Yang, Ruihan and Xu, Huazhe and Wu, Yi and Wang, Xiaolong},
  booktitle={Advances in Neural Information Processing Systems},
  year={2020}
}

@article{ren2021probabilistic,
  title={Probabilistic Mixture-of-Experts for Efficient Deep Reinforcement Learning},
  author={Ren, Jie and Li, Yewen and Ding, Zihan and Pan, Wei and Dong, Hao},
  journal={arXiv preprint arXiv:2104.09122},
  year={2021}
}

@inproceedings{haarnoja2018soft,
  title={Soft actor-critic: Off-policy maximum entropy deep reinforcement learning with a stochastic actor},
  author={Haarnoja, Tuomas and Zhou, Aurick and Abbeel, Pieter and Levine, Sergey},
  booktitle={International conference on machine learning},
  pages={1861--1870},
  year={2018},
  organization={PMLR}
}

@article{han2025kungfubot2,
  title={{KUNGFUBOT2}: Learning Versatile Motion Skills for Humanoid Whole-Body Control},
  author={Han, Jinrui and Xie, Weiji and Zheng, Jiakun and Shi, Jiyuan and Zhang, Weinan and Xiao, Ting and Bai, Chenjia},
  journal={arXiv preprint arXiv:2509.16638},
  year={2025}
}

@article{ji2024exbody2,
  title={ExBody2: Advanced Expressive Humanoid Whole-Body Control},
  author={Ji, Mazeyu and Peng, Xuanbin and Liu, Fangchen and Li, Jialong and Yang, Ge and Cheng, Xuxin and Wang, Xiaolong},
  journal={arXiv preprint arXiv:2412.13196},
  year={2024}
}

@article{chen2025gmt,
  title={GMT: General Motion Tracking for Humanoid Whole-Body Control},
  author={Chen, Zixuan and Ji, Mazeyu and Cheng, Xuxin and Peng, Xuanbin and Peng, Xue Bin and Wang, Xiaolong},
  journal={arXiv preprint arXiv:2506.14770},
  year={2025}
}

@article{zakka2026mjlab,
  title={mjlab: A Lightweight Framework for GPU-Accelerated Robot Learning},
  author={Zakka, Kevin and Liao, Qiayuan and Yi, Brent and Lay, Louis Le and Sreenath, Koushil and Abbeel, Pieter},
  journal={arXiv preprint arXiv:2601.22074},
  year={2026}
}

@article{harvey2020robust,
  title={Robust motion in-betweening},
  author={Harvey, F{\'e}lix G and Yurick, Mike and Nowrouzezahrai, Derek and Pal, Christopher},
  journal={ACM Transactions on Graphics (TOG)},
  volume={39},
  number={4},
  pages={60:1--60:12},
  year={2020},
  publisher={ACM New York, NY, USA}
}

@article{van2008visualizing,
  title={Visualizing data using t-SNE.},
  author={Van der Maaten, Laurens and Hinton, Geoffrey},
  journal={Journal of machine learning research},
  volume={9},
  number={11},
  year={2008}
}

@misc{amazonfar2025holosoma,
  author       = {{Amazon Frontier AI \& Robotics (FAR)}},
  title        = {Holosoma: An Open-Source Framework for Humanoid Robot Learning},
  howpublished = {\url{https://github.com/amazon-far/holosoma}},
  year         = {2025}
}

\end{document}